\documentclass{l4dc2024}

\title[PRFT]{Adapting Image-based RL Policies via Predicted Rewards}
\usepackage{times}

\usepackage{caption}

\usepackage{booktabs}       % professional-quality tables
\usepackage{amsfonts}       % blackboard math symbols
\usepackage{nicefrac}       % compact symbols for 1/2, etc.
\usepackage{microtype}      % microtypography
\usepackage{natbib}
\usepackage{graphicx}

\usepackage{algorithm}
\usepackage{algorithmic}

\usepackage{amsmath, amssymb}
\usepackage{wrapfig}
\usepackage{cleveref}

\usepackage{wrapfig}
\makeatletter
\def\BState{\State\hskip-\ALG@thistlm}

\newcommand{\cs}{\mathbf{s_t}} 

\newcommand{\ca}{\mathbf{a_t}} 
\newcommand{\gH}{\mathcal{H}}
\newcommand{\gR}{\mathcal{R}}
\newcommand{\gA}{\mathcal{A}}
\newcommand{\E}{\mathbb{E}}

\coltauthor{\Name{Weiyao Wang} \Email{wwang121@jhu.edu}\\
 \Name{Xinyuan Fang} \Email{xfang22@jhu.edu}\\
 \Name{Gregory D. Hager} \Email{hager@cs.jhu.edu}\\
 \addr Computer Science Department, Johns Hopkins University}

\begin{document}

\maketitle

\begin{abstract}%
Image-based reinforcement learning (RL) faces significant challenges in generalization when the visual environment undergoes substantial changes between training and deployment. Under such circumstances, learned policies may not perform well leading to degraded results. Previous approaches to this problem have largely focused on broadening the training observation distribution, employing techniques like data augmentation and domain randomization. However, given the sequential nature of the RL decision-making problem, it is often the case that residual errors are propagated by the learned policy model and accumulate throughout the trajectory, resulting in highly degraded performance. 
In this paper, we leverage the observation that predicted rewards under domain shift, even though imperfect, can still be a useful signal to guide fine-tuning. We exploit this property to fine-tune a policy using reward prediction in the target domain. We have found that, even under significant domain shift, the predicted reward can still provide meaningful signal and fine-tuning substantially improves the original policy. Our approach, termed Predicted Reward Fine-tuning (PRFT), improves performance across diverse tasks in both simulated benchmarks and real-world experiments. More information is available at project web page: \href{https://sites.google.com/view/prft}{https://sites.google.com/view/prft}.
\end{abstract}


\section{Introduction}
Image-based reinforcement learning (RL) has gained substantial attention with success in gaming~\citep{mnih2013playing}, robotic manipulation~\citep{amarjyoti2017deep}, and autonomous driving~\citep{chen2021interpretable}, among other areas. However, the visual environment often undergoes significant appearance changes, such as lighting, textures, and camera poses, between the training and testing phases. This leads to performance degradation due to the well recognized challenge of input domain gap~\citep{zhao2020sim}. This generalization challenge is particularly acute in sequential decision-making processes such as RL where deviations at each step can accumulate, significantly exacerbating performance degradation during rollout~\citep{wang2019generalization}.

\begin{figure}[t]
\centering
\includegraphics[width=5in]{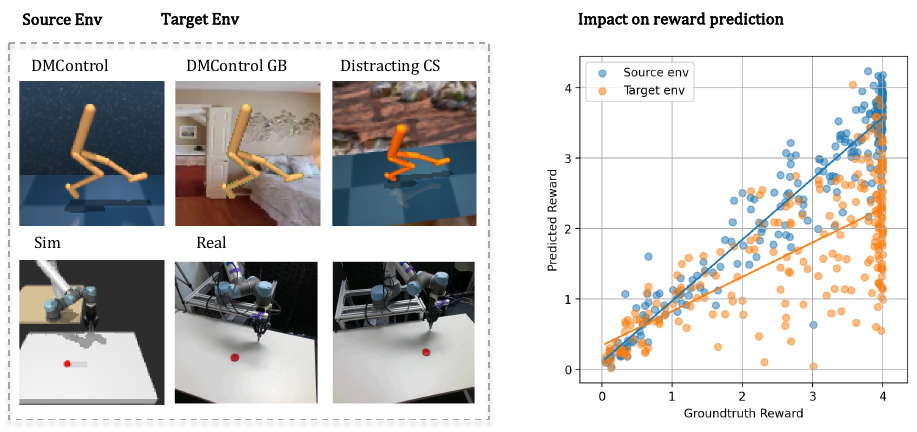}
\caption{\textbf{Left:} Example of source environment observations and target environment observations. \textbf{Right:} Illustration of domain shift effect on reward prediction. Samples are collected using a trained policy on the source walker walk environment, and the domain shift effect is tested by evaluating predicted rewards under both the source and the target environment (\texttt{video\_hard} in DMControl GB) with the same underlying states. Fitted linear regression for the predicted rewards against the groundtruth rewards for both source and target environment are plotted for visualization.}
\label{fig:shift_effect}
\end{figure}

To improve generalization across different domains, a common approach 
involves domain randomization~\citep{ganin2016domain, chen2021understanding} or data augmentation~\citep{ma2022comprehensive,kirk2021survey}. These methods broaden the training observation distribution with the intention of covering the target environment encountered during testing. However, anticipating domain shift can be challenging, rendering this method less effective, particularly when the domain shift is significant. Another method to correct errors under visual domain shift is to fine-tune under the target deployment domain~\citep{wang2018deep, guo2019spottune}. This, however, is often impractical for robot learning problems as, in many cases, rewards are specified using internal states only available in the training domain (e.g. in simulation). 

To address this issue, we investigate an alternative approach: fine-tuning the policy in the reward-free target environment using predicted rewards from observations. The proposed method stems from the observation that domain shifts impact the policy model and the reward prediction model in distinct ways. 
\begin{itemize}
    \item RL is able to effectively refine a policy in the presence of random reward space noise~\citep{sun2021reward, wang2020reinforcement, eysenbach2021maximum}, which makes improvement with imperfect rewards feasible.
    Meanwhile, error in action will accumulate throughout the rollout, leading to significantly deteriorated episodic performance with~\citep{ross2010efficient}. The regret can grow quadratically with the number of timestep and it is difficult to address.
    
    \item  Not all of the domain shift impact on reward prediction will alter the induced optimal policy. As depicted in Figure~\ref{fig:shift_effect}, aside from exhibiting larger errors, reward prediction in the target environment is become more conservative for out-of-distribution samples and can be regarded as undergoing a linear transformation in the form of $\hat{r}= k r + b$ with $k > 0, b \in \mathcal{R}$. Such linear transformation is known to maintain the optimal policy.

    This suggests that part of the reward prediction error can be considered as benign. In contrast, errors in action do not possess similar property, rendering them less forgiving.
\end{itemize}

Building on these insights, we propose to jointly learn (at training time) a policy and a reward prediction model. The policy is then fine-tuned using the predicted rewards in the target testing environment prior to testing and deployment. Through extensive experiments in both simulation and real-world experiments, we show that the reward prediction model generalizes well across visual effect shifts and significantly enhances policy performance through fine-tuning. 
We benchmark our approach against others addressing the domain adaptation problem in RL, including methods that incorporate data augmentation~\citep{hansen2021stabilizing} and self-supervised test-time training~\citep{hansen2020deployment} for domain adaptation. Our experimental results demonstrate that our approach outperforms the baseline methods by substantial margins across various benchmark environments.

In summary, we propose a novel approach for image-based RL that employs a reward prediction function to adapt to visual domain shifts in the testing environment. The efficacy of this method is built upon the structural advantage of the reward model in generalizing RL policies. Our approach presents a promising direction to the challenge of domain adaptation in image-based RL.

\section{Related Work}
Reinforcement learning that leverages visual input has recently gathered substantial interest due to its broad range of real-world applications. Techniques that uses learned image encoders (e.g., convolutional neural networks) to reinforcement learning algorithms have demonstrated impressive achievements by learning policies directly from pixel images~\citep{finn2016deep,lin2019adaptive, lyle2021effect, he2022reinforcement}.

Various approaches have been proposed to address the generalization problem in visual RL. One common category involves domain randomization~\citep{ganin2016domain, chen2021understanding} or data augmentation during training~\citep{kostrikov2020image, laskin2020reinforcement, ma2022comprehensive}. One challenge with these methods is that an expanded training observation distribution complicates the optimization of RL policies. To mitigate this issue, a line of research~\citep{hansen2021generalization,fan2021secant} proposed to decouple augmentation from policy learning by latent space regularization or policy distillation. ~\cite{hansen2021stabilizing} further identified that direct application of regularization introduces non-deterministic Q-targets and over-regularization, leading to inefficiency in policy optimization. Their method, SVEA, proposes to jointly optimize the Q-function with both augmented and non-augmented data to improve stability and sample efficiency.

Instead of trying to be invariant to all domains, some works also aim to adapt policy to some specific target environment without reward access, which shares the same setting as our method. One approach uses generative adversarial networks (GANs)~\citep{goodfellow2020generative} to translate images~\citep{zhang2019vr, rao2020rl,ho2021retinagan} or latent features~\citep{yoneda2021invariance} from target domains to source domain and then feed into policies. Despite being effective in some cases, this line of research requires access to a large amount of target domain observations to capture the observational space distribution. Moreover, the training can be challenging with generative adversarial training.
Another approach under this setting is to perform adaptation through self-supervised auxillary task under the target environment. PAD~\citep{hansen2020deployment} jointly learns an inverse dynamics model (IDM) alongside RL learning during training. In testing, it fine-tunes the image encoder by optimizing the IDM objective to adapt to the target environment. Fine-tuning on this self-supervised task circumvents the problem of reward signal unavailability in the test deployment environment. However, its benefits are limited since this signal adapts to the new transition dynamics but is relatively indirect from the task to be performed.

In a similar vein to our method, other approaches explore the use of an evaluation model to provide feedback to fine-tune policies. PAFF~\citep{ge2022self} leverages a robust vision language foundation model to label executed instruction-conditioned policies under testing domain with descriptions and then fine-tunes policies with generated descriptions using imitation learning (IL). More broadly in the space of Natural Language Processing (NLP), the recent popular method Reinforcement Learning from Human Feedback (RLHF)~\citep{ouyang2022training} first learns a reward model based on human preferences and then fine-tunes the language generative model based on the learned reward model. Meanwhile in this paper, we systematically examine the benefit of fine-tuning with a reward model in the context of visual domain adaptation for RL policies.

\section{Method}

Formally, we frame our problem as adaptation of a visual policy. Consider a Markov decision process (MDP)~\citep{sutton2018reinforcement}
$\mathcal{M} = \{\mathcal{S},\mathcal{A},\mathcal{P},\mathcal{R},\gamma\}$
where $S$ and $A$ are the state and action spaces respectively. $P: S \times A \rightarrow S$ is the transition function, $R: S \times A \rightarrow \mathrm{R}$ represents the scalar reward, and $\gamma$ represents the discount factor. We assume that the agent cannot directly observe state space $S$ but only receives higher-dimensional input $O$ (e.g., pixel images) as observations. Let $\pi(o,a) \in \Pi: O,A \rightarrow [0,1]$ be the policy that maps observation $o\in O$ and action $a$ to probability. The source environment and target environment differ only in the mapping of the same state $S$ to different observations $O$ due to visual appearance shifts, while all other elements remain the same. Moreover, we presume that the agent can access the target environment to gather interactions but without groundtruth reward provided.

We first consider the standard RL framework, where an agent interacts with an environment over a sequence of discrete time steps. At each time step $t$, the agent receives an observation $o_t$ from observation space $O$, takes an action $a_t$ from a set of possible actions $\mathcal{A}$, and receives a scalar reward $r_t$. The goal is to learn a policy $\pi$ that maximizes the expected cumulative reward, defined as:

\begin{equation}
J(\pi;r) \triangleq \E_{\pi} \bigg[ \sum_{t=1}^T r(\cs, \ca)\gamma^t \bigg],
\end{equation}
where $T$ represents the termination timestep.

\begin{figure*}[t]
\centering
\includegraphics[width=5.25in]{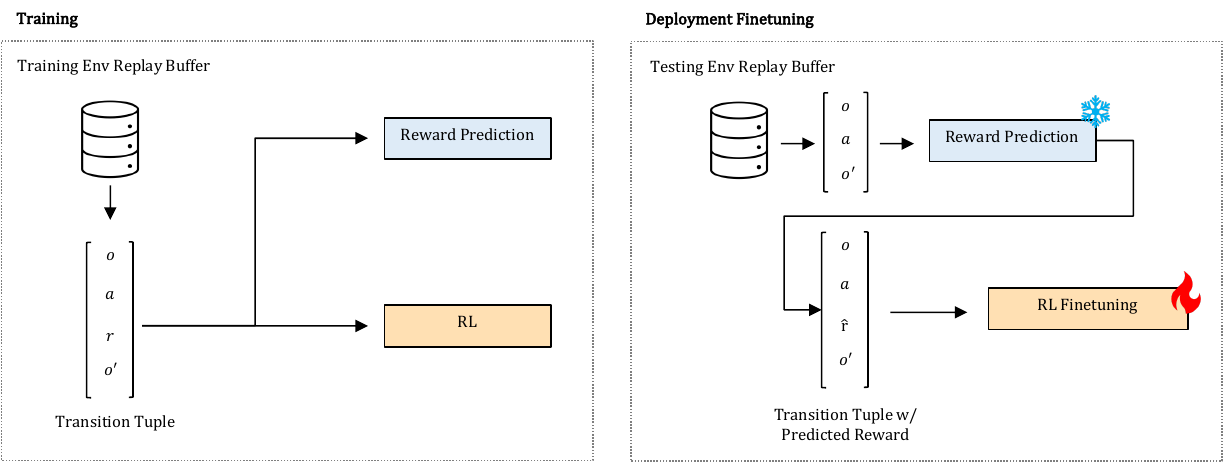}

\caption{\textbf{Left:} During training, we optimize the reward prediction module along with reinforcement learning using sampled transition tuples from replay buffer. \textbf{Right:} During deployment finetuning, we use the transition tuples with predicted reward to finetune the reinforcement learning policy. The reward prediction module is frozen in this stage.}
\label{architecture}
\end{figure*}

We denote the reward prediction function as $\phi(o,a)$. This function takes in an observation $o$ and an action $a$, and then it outputs a scalar reward prediction $\hat{r}$. A neural network is utilized to model this function, which we denote as $\phi$ with parameters $\theta_r$. We train this neural network to minimize the discrepancy between the predicted reward $\hat{r}$ and the actual reward $r$. This discrepancy is measured by the mean squared error, defined as follows:

\begin{equation}
\mathcal{L}_{\phi(\theta_r)} = \mathbb{E}_{(o,a,r) \sim \mathcal{B}_\text{train}}[(\hat{r} - r)^2] = \mathbb{E}_{(o,a,r) \sim \mathcal{B}_\text{train}}[(\phi(o,a; \theta_r) - r)^2].
\end{equation}
In this equation, $\mathcal{B}_\text{train}$ represents the training replay  dataset of observed states and rewards.

\begin{algorithm}
\caption{Training and Fine-tuning PRFT}
\label{alg:prft}
\begin{algorithmic}[1] % The number tells where the line numbering should start

\STATE \textbf{Train:}
\STATE \textit{Inputs:} Reward prediction function $\phi$, its parameters $\theta_r$, policy $\pi$, training replay buffer $\mathcal{B}_\text{train}$
\FOR{each epoch}
    \STATE \quad Execute $\pi$ in the environment to get $(o, a, r, o')$, store in $\mathcal{B}_\text{train}$
    \STATE \quad Sample $(o, a, r, o')$ from $\mathcal{B}_\text{train}$
    \STATE \quad Update $\pi$ using MaxEnt RL algorithm with reward $(o, a, r, o')$
    \STATE \quad Update $\theta_r$ to minimize $\mathcal{L}_{\phi(\theta_r)} = (\phi(o,a; \theta_r) - r)^2$
\ENDFOR

\STATE 

\STATE \textbf{Fine-tune:}
\STATE \textit{Inputs:} Reward prediction function $\phi$, its parameters $\theta_r$, policy $\pi$, replay buffer $\mathcal{B}_{ft}$
\STATE \quad Freeze $\phi$
\FOR{each epoch}
    \STATE \quad Execute $\pi$ in the environment to get $(o, a, o')$
    \STATE \quad Compute $\hat{r}$ using $\phi(o, a; \theta_r)$, store $(o, a, \hat{r}, o')$ in $\mathcal{B}_{ft}$
    \STATE \quad Update $\pi$ using MaxEnt RL algorithm with samples from $\mathcal{B}_{ft}$
\ENDFOR

\end{algorithmic}
\end{algorithm}

PRFT is outlined in Algorithm~\ref{alg:prft} with the process of training and fine-tuning. 
During the training phase, a policy and reward prediction function are jointly learned. After training, the reward prediction function is frozen and the policy is fine-tuned. During the fine-tuning stage, the agent interacts with the environment and the experiences are stored in a replay buffer. The stored experiences are then used to compute reward predictions which in turn guide the fine-tuning of the policy. This ensures a better alignment of the policy to perform well in the target environment.

We employ MaxEnt maximum entropy (MaxEnt) RL algorithm to optimize policy $\pi$ during both the training and fine-tuning phases.
Recall that the original reward function is defined as $J_{\text(\pi; r)} \triangleq \E_{\pi} \bigg[ \sum_{t=1}^T r(\cs, \ca)\gamma^t \bigg]$. The MaxEnt RL objective is to maximize the sum of the expected reward and the conditional action entropy, represented by
$J_\text{MaxEnt}(\pi; r) \triangleq \E_{\pi} \bigg[ \sum_{t=1}^T r(\cs, \ca)\gamma^t + \alpha \gH_\pi[\ca \mid \cs] \bigg]$, where $\gH_\pi[\ca \mid \cs] = \int_{\gA} \pi(\ca \mid \cs) \log \frac{1}{\pi(\ca \mid \cs)} d\ca$ denotes the entropy of the action distribution. The \emph{entropy coefficient} $\alpha$ balances the reward and entropy terms. %; for our analysis, we set $\alpha = 1$ as in~\cite{eysenbach2021maximum}.

We choose to use MaxEnt RL in particular because it is robust to some degree of misspecification in the reward function.
\cite{eysenbach2021maximum} shows that assuming $\alpha = 1$, the reward function is finite and the policy has support everywhere (i.e., $\pi(\ca \mid \cs) > 0$ for all states and actions), there exists a positive constant $\epsilon>0$ such that optimizing the MaxEnt RL objective $J_\text{MaxEnt}(\pi, \hat{r})$ is equivalent to optimizing a lower bound of the objective function $J(\pi, \tilde{r})$:

\begin{equation*}
 \min_{\tilde{r} \in \tilde{\gR}(\pi)} \E \Big[\sum_t \tilde{r}(\cs, \ca) \Big] = J_\text{MaxEnt}(\pi; p, r) \quad \forall \pi,
\end{equation*}
where the adversary chooses a reward function from the robust set
\begin{equation}
     \tilde{\gR}(\pi) \triangleq \left\{ \tilde{r}(\cs, \ca) \; \bigg \vert \; \E_\pi \Big[\sum_t \log \int_{\gA} \exp(r(\cs, \ca') - \tilde{r}(\cs, \ca')) d\ca'\Big] \le \epsilon \right\}. \label{eq:robust-set-rewards}
\end{equation}

Therefore applying MaxEnt RL to one reward function, $r(\cs, \ca)$, results in a policy that is guaranteed to also achieve high return on a range of other reward functions, $\tilde{r}(\cs, \ca) \in \tilde{\gR}$.
This suggests that we can employ MaxEnt RL algorithm such as SAC~\citep{haarnoja2018soft} to improve a policy's performance on the unknown true reward objective even if the provided reward is imperfect (i.e., from a reward predictor under domain shift). With the help of the reward prediction and MaxEnt RL algorithm, we are able to fine-tune the policy on test domain without groundtruth reward signal provided. In the following section, we are going to show that such fine-tuning stage significantly improves the policy performance on various simulated and real-robot tasks.

\section{Experiments}

We aim to understand the impact of domain adaptation on an agent’s ability to generalize out-of-distribution. This section compares PRFT with a state-of-the-art baseline in image-based RL: DrQ~\citep{laskin2020reinforcement} and one that uses data-augmentation: SVEA~\citep{hansen2021stabilizing}. As discussed in the related work, this method enhances generalization capabilities by expanding the support of the training distribution. 
% We hypothesize that the invariance produced this way is less effective than fine-tuning with predictions at test time, which provides task-related signals on specific instances of perceptual variation. 
We use SVEA as a data augmentation baseline to supplement our method, denoting as SVEA + PRFT. In our experiments, we also compare against SVEA+IDM, the combination of SVEA and inverse dynamics model, and SVEA+PAD, the combination of SVEA and policy adaptation during deployment~\citep{hansen2020deployment} which is a baseline method that adapts the policy through self-supervised auxiliary tasks.
All methods in comparison employs MaxEnt RL algorithm SAC~\citep{haarnoja2018soft} as base learning algorithm.

\textbf{Implementation:} For all methods, we adopt the same network architecture as~\cite{hansen2021stabilizing}: a 11-layer ConvNet followed by 3-layer MLP with 1024 hidden units. We use the \textit{random overlay} data augmentation~\citep{hansen2021generalization} as a part of SVEA training. \textit{random overlay} interpolates observation image with a random chosen image from Places dataset~\citep{zhou2017places}.

\begin{figure}[]
\centering
\includegraphics[width=5.25in]{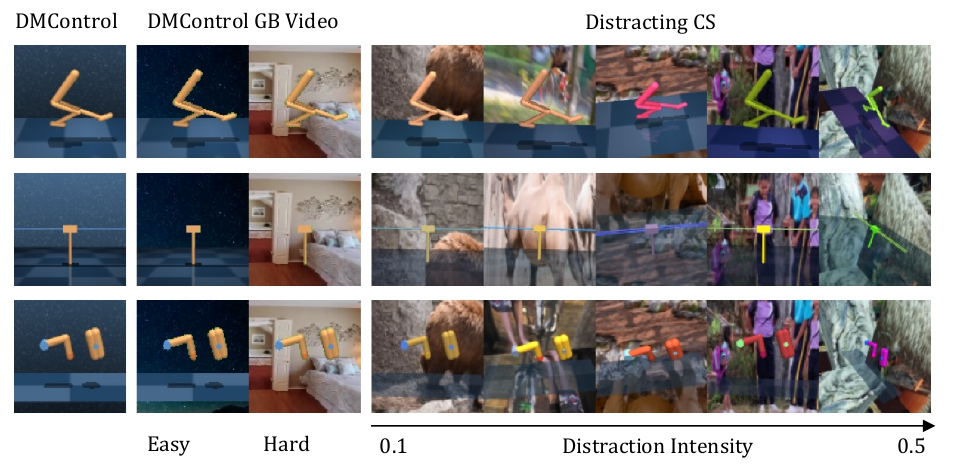}

\caption{Samples from deepmind control suite (DMControl), deepmind control generalization benchmark (DMControl GB) with video background (easy and hard), and distracting control suite (Distracting CS) with intensity from 0.1 to 0.5.}
\label{vis_sample}
\end{figure}

\subsection{PRFT on Simulated Environments}

We conduct experiments on 6 domains from the DeepMind Control Suite (DMControl~\citep{tassa2018deepmind}) and treat it as the source domain for training the RL agents. For the target domain, we use 1) the DMControl with video background introduced in DMControl Generalization Benchmark (DMControl GB)~\citep{hansen2021stabilizing} and 2) Distracting Control Suite (Distracting CS)\cite{stone2021distracting}. Distracting CS adds background, color, and camera pose distractions, making it an ideal benchmark for testing the generalization and robustness of reinforcement learning algorithms. All simulated experiments are performed across 4 random seeds with 20 episodes during evaluation. 

% \newpage
We first present the evaluation results on DMControl GB with \texttt{video\_easy} and \texttt{video\_hard} domains in Table~\ref{tab:dmcgb}. The table compares different methods, including DrQ~\citep{kostrikov2020image}, SVEA~\citep{hansen2021stabilizing}, PAD~\citep{hansen2020deployment}, and our method, PRFT, based on episodic reward. Notably, in 10 out of 12 tasks, our proposed PRFT performs the best, surpassing competing approaches. By fine-tuning with predicted reward, we observe a clear improvement in reward across different tasks. This highlights the effectiveness of PRFT in adapting the policy to the target domain and enhancing performance. We notice that in \texttt{video\_hard} version of \texttt{walker\_run} environment, our method brought negative change to the performance. This suggests that the error in reward prediction could be too large and make it a harmful fine-tuning signal in this case.

\begin{table}[]
\centering
\scriptsize
\begin{tabular}{@{}lccccc@{}}
\toprule
\textbf{Task (\texttt{video\_easy})} & {DrQ} & {SVEA} & {+IDM} & {+PAD} & {+PRFT (ours)} \\
\midrule
\texttt{finger\_spin} & 477 $\pm$ 435 & 770 $\pm$ 306 & 661 $\pm$ 450 & 685 $\pm$ 463 & \textbf{944 $\pm$ 37} \\
\texttt{cartpole\_balance} & 717 $\pm$ 162 & 862 $\pm$ 128 & 640 $\pm$ 269 & 646 $\pm$ 296 & \textbf{897 $\pm$ 114} \\
\texttt{cartpole\_swingup} & 496 $\pm$ 409 & 825 $\pm$ 28 & 644 $\pm$ 298 & 644 $\pm$ 295 & \textbf{826 $\pm$ 11} \\
\texttt{walker\_stand} & 890 $\pm$ 109 & \textbf{940 $\pm$ 17} & 904 $\pm$ 102 & 899 $\pm$ 110 & 936 $\pm$ 15 \\
\texttt{walker\_walk} & 630 $\pm$ 258 & 843 $\pm$ 141 & 870 $\pm$ 82 & 867 $\pm$ 42 & \textbf{878 $\pm$ 43} \\
\texttt{walker\_run} & 183 $\pm$ 102 & 240 $\pm$ 34 & 243 $\pm$ 21 & 244 $\pm$ 21 & \textbf{258 $\pm$ 17} \\ \hline
\textbf{Task (\texttt{video\_hard})} & & & & & \\\hline
\texttt{finger\_spin} & 16 $\pm$ 27 & 151 $\pm$ 118 & 107 $\pm$ 75 & 97 $\pm$ 67 & \textbf{193 $\pm$ 115} \\
\texttt{cartpole\_balance} & 214 $\pm$ 30 & 303 $\pm$ 64 & 244 $\pm$ 23 & 254 $\pm$ 28 & \textbf{860 $\pm$ 198} \\
\texttt{cartpole\_swingup} & 169 $\pm$ 33 & 376 $\pm$ 70 & 189 $\pm$ 45 & 181 $\pm$ 54 & \textbf{510 $\pm$ 289} \\
\texttt{walker\_stand} & 287 $\pm$ 103 & 849 $\pm$ 64 & 764 $\pm$ 183 & 769 $\pm$ 176 & \textbf{904 $\pm$ 47} \\
\texttt{walker\_walk} & 105 $\pm$ 75 & 377 $\pm$ 172 & 556 $\pm$ 199 & 541 $\pm$ 212 & \textbf{560 $\pm$ 114} \\
\texttt{walker\_run} & 36 $\pm$ 9 & 171 $\pm$ 40 & 173 $\pm$ 33 & \textbf{174 $\pm$ 36} & 154 $\pm$ 45 \\
\bottomrule
\end{tabular}
\caption{Evaluation by episodic cumulative rewards (mean ± standard deviation) for 6 DMControl GB environments. Comparison of methods includes DrQ, SVEA, SVEA+IDM, SVEA+PAD, and our method SVEA+PRFT. SVEA is omitted from the method name of the last three columns in the chart for better readability.}
\label{tab:dmcgb}
\end{table}

\begin{figure}[t]
\centering
\includegraphics[width=6in]{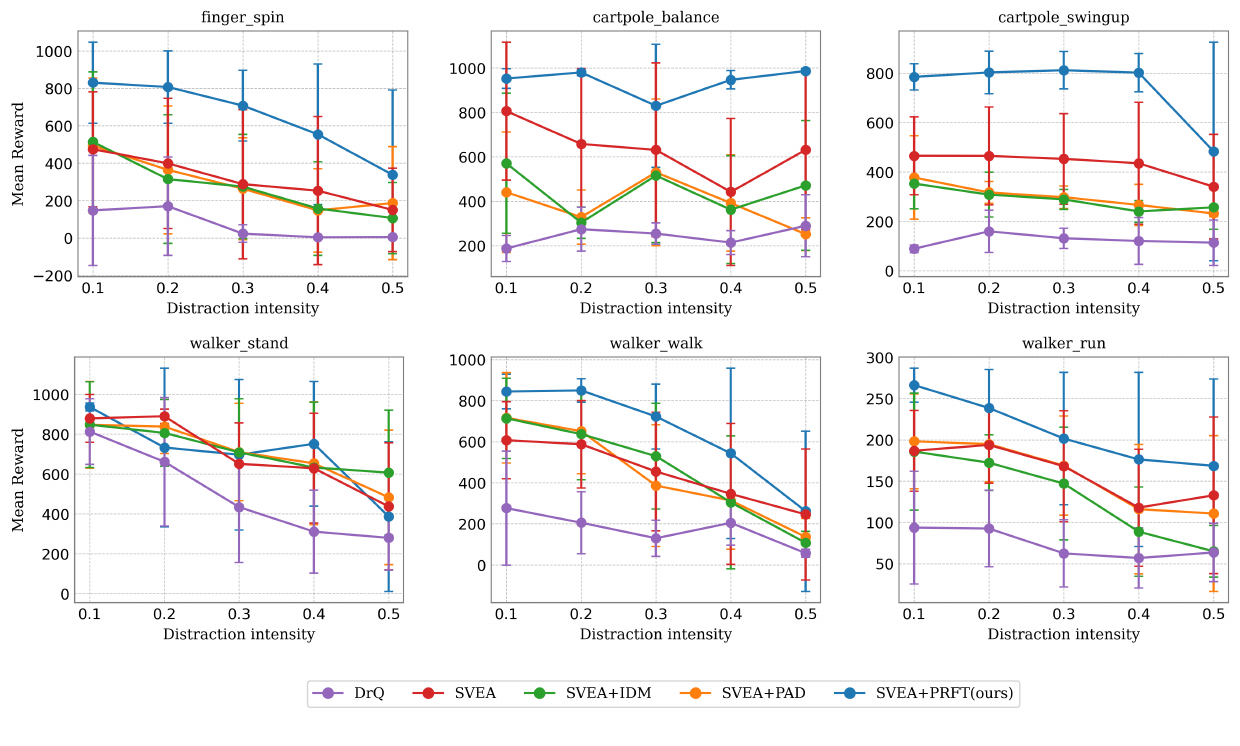}

\caption{Evaluation in environments under distracting control suite with varying degrees of distraction intensities. Our method significantly outperforms baseline methods in five out of six environments. Error bar shows one standard deviation.}
\label{dcs}
\end{figure}

Beyond domain shifts in the background, we further tested our method in the Distracting Control Suite, which additionally introduces distractions in color and camera pose with controlled intensities ranging from 0.1 to 0.5. Figure~\ref{dcs} visualizes the performance of the different methods, as a function of the intensity of the distractions. Since the baseline method SVEA is trained with image augmentation, it does exhibit some robustness to distraction. However, we see this robustness rapidly diminishes as the distraction intensity increases. In particular, large changes to camera pose or the image background proved challenging for standard augmentation procedures. Comparatively, PRFT makes it much smoother and slower degradation of performance for most of the environments. The improvement is particularly prominent under high distraction intensity, highlighting the robustness of our reward prediction in generating useful signals despite heavy domain shifts. This supports our hypothesis that adaptation powered by predicted rewards can significantly improve the target environment performance abilities of the policy.

\begin{wrapfigure}{r}{0.5\linewidth}
\centering
\includegraphics[width=0.81\linewidth]{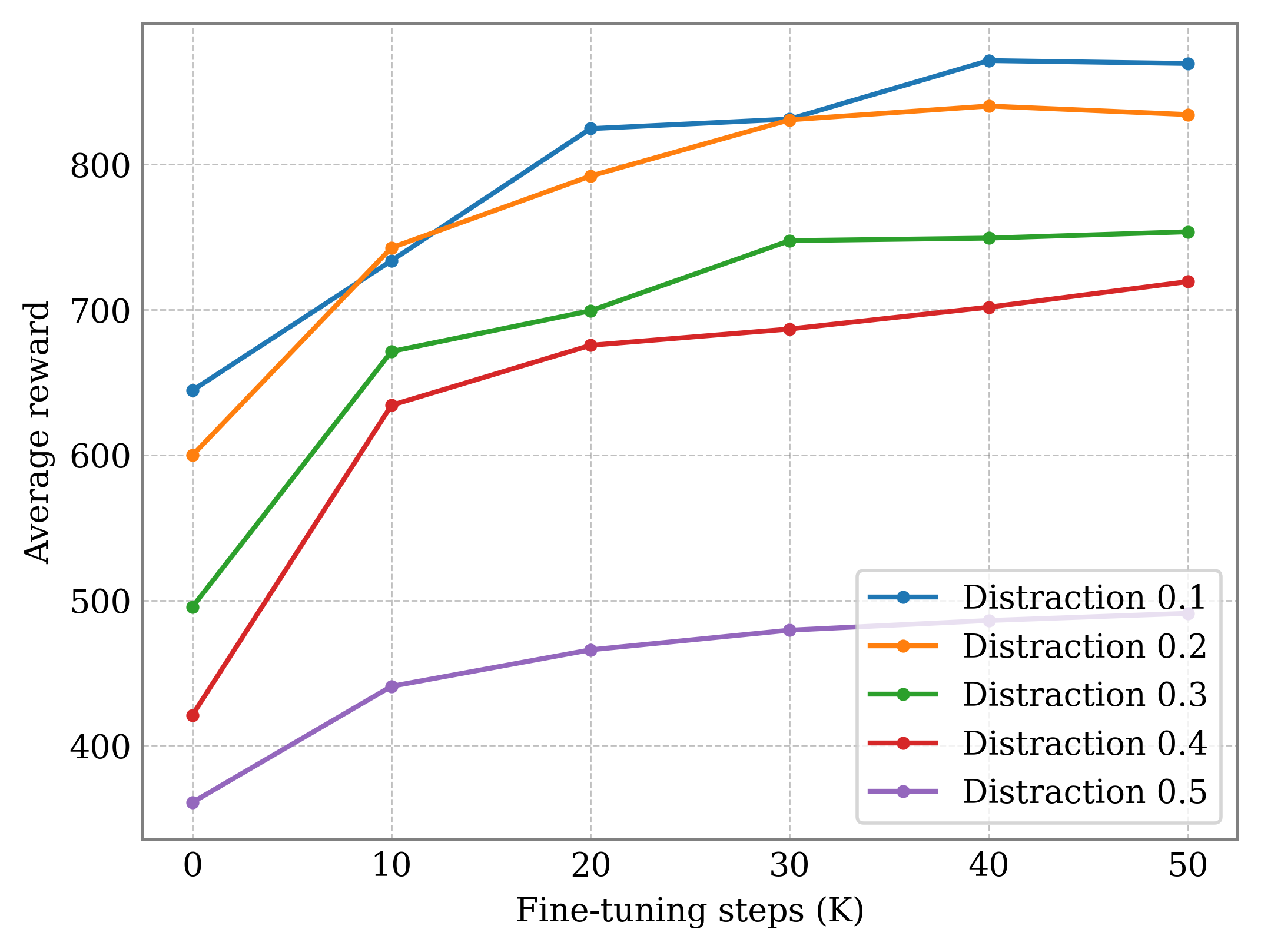}
\includegraphics[width=0.8\linewidth]{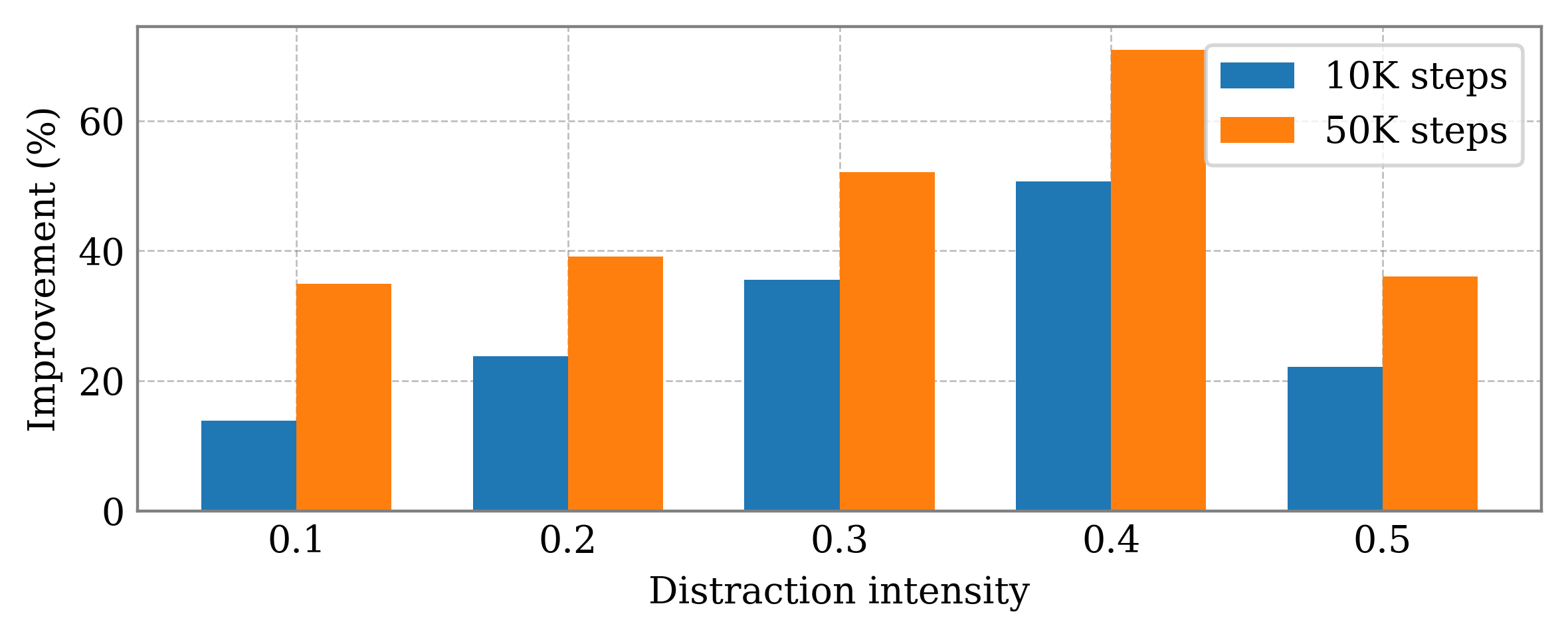}
\caption{\textbf{Top:} Policies improve during fine-tuning using predicted rewards. 
Average episodic rewards over four tasks and four random seeds are plotted. 
\textbf{Bottom:} Relative improvement of average rewards across different distraction intensities
at 10K and 50K fine-tuning steps.
}
\label{fig:prft_improvement}
\end{wrapfigure}

To better visualize the improvements in rewards brought about by fine-tuning, we plotted the average reward improvements for each distraction intensity in Figure~\ref{fig:prft_improvement}. As depicted in the graphs, our method, PRFT, enhances the average rewards across all distraction intensities swiftly, using as few as 10K steps. The performances then continue to increase and converge at 50K steps. On comparing the improvements across different distraction intensities, we found that the largest improvement was made at a moderately high distraction intensity of 0.4. Here, fine-tuning with 10K steps brings about a 50\% improvement, and fine-tuning with 50K steps brings about a 70\% improvement. The improvements at lower and higher distraction intensities were smaller but still substantial. For small distraction values of 0.1 and 0.2, this is because the zero-shot transfer already performs well, thereby limiting the scope for improvement. For large distraction values of 0.5, the accuracy of the predicted reward also drops, which results in diminishing benefits from using it as a fine-tuning signal.

\begin{wraptable}{r}{0.4\textwidth}
  \centering
  % \scriptsize
  \label{tab:success-rates}
  \begin{tabular}{ll}
    \toprule
    Method & Success Rate \\
    \midrule
    DrQ & 0.32 \\
    SVEA & 0.52 \\
    +PAD & 0.48 \\
    +PRFT (ours) & \textbf{0.68} \\
    \bottomrule
  \end{tabular}
  \caption{Evaluation on sim-to-real.}
\end{wraptable}

\subsection{Sim-to-real Transfer}
\begin{figure}[t]
\centering
\includegraphics[width=6in]{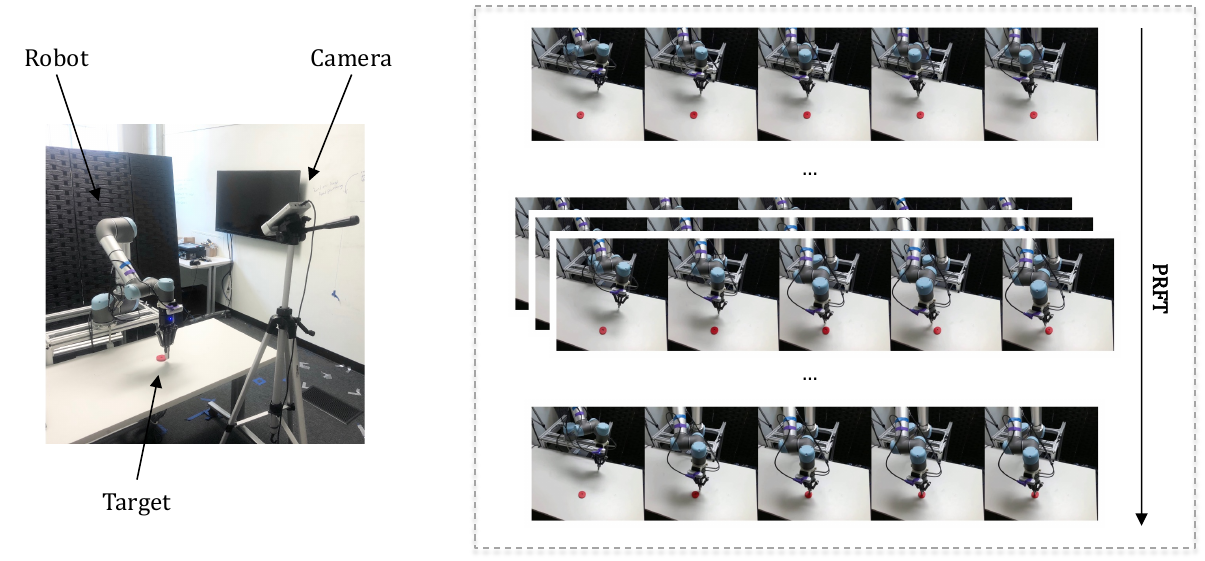}

\caption{\textbf{Left:} Our real-world experiment setup with robot, camera and target. \textbf{Right:} Example rollouts of progress made in real-world predicted reward fine-tuning. The robot is able to reach target that is not achievable before fine-tuning.}
\label{vis_sample}
\end{figure}
We are also interested in the ability of PRFT to bridge the gap between simulated and real world robotics environments. To do so we define a reaching task where a target position is given by a red disc placed on a table. The agent’s objective is to controls the arm so that the end-effector reaches this target location. Our goal is to train a policy in simulation, and then transfer the policy to a real UR-5 robot at test time. The same as with previous experiments, the test time agent receives no rewards. In both simulation and the real world, the policy’s only input is an image from a camera placed in front of the robot and table. The action space is a 2D position controller that drives a small movement of the robotic gripper.

We carry out the experiment by first training a policy with DrQ and SVEA [14] in simulation. For adaptation, we apply PAD and PRFT to finetune the SVEA policy in the real environment to bridge the sim-to-real gap. We evaluated the success rate of zero-shot and the adapted policies over 25 real episodes. We consider the episode to be success if the gripper’s tip is within 5 cm of the center of the target.
Due to the challenging domain shift between simulator and real world, both the zero-shot policies and PAD fails to adapt adequately. One of the failure cases is repeating the same action of moving the gripper to an edge of the table, regardless of the given goal location. This results in a final success rate around 50\% for SVEA on zero-shot. On the same task, PRFT is able to adapt to this domain shift, achieving a final success rate around 70\%.

In summary, our results demonstrate the robustness of our method in handling domain shift introduced in both simulated and real environments. The substantial improvements achieved by our approach, even under high distraction intensity, highlight the effectiveness predicted rewards as fine-tuning signals.

\section{Conclusion}

In this paper we have presented and evaluated a novel approach, PRFT (Predicted Reward Fine-Tuning), for adapting policy under domain shift. Our method demonstrated superior performance across various robotic environments, outperforming baseline methods in large margins. Notably, our approach exhibited significant improvement even under high distraction intensity, highlighting the robustness of our method under large domain shifts.

Our results indicate that, imperfect reward prediction is still a useful fine-tuning signal in the test domain. This, however, no longer holds when the error exceeds certain level. When large domain shifts happens, incorrectly predicted reward may misguide the policy fine-tuning, leading to even worse performance compared to zero-shot testing. How to detect such scenarios and mitigate accordingly is one direction for future work.

\acks{This work was supported by the U.S. National Science Foundation grants IIS-1900952 to Johns Hopkins University.}

\bibliography{bib}

\end{document}